\documentclass[sigconf]{acmart}
\AtBeginDocument{%
  }

\usepackage{booktabs}
\usepackage{siunitx}
\usepackage{amsfonts}
\usepackage{multirow}
\usepackage{color}
\usepackage{bm}
\usepackage{amsmath, amsthm}

\usepackage{graphicx}
\usepackage{caption}
\usepackage{subcaption}
\usepackage{xcolor}
\usepackage{url}
\usepackage{colortbl} 
\usepackage{algorithm}
\usepackage{algpseudocode}
\usepackage{balance}
\begin{document}

\title[Turing Patterns for Multimedia: Reaction-Diffusion Multi-Modal Fusion \\ for Language-Guided Video Moment Retrieval]{Turing Patterns for Multimedia: Reaction-Diffusion Multi-Modal Fusion for Language-Guided Video Moment Retrieval}
\author{Xiang Fang}
\email{xfang9508@gmail.com}
\author{Wanlong Fang}
\email{wanlongfang@gmail.com}
\affiliation{%
  \institution{Nanyang Technological University}
   \city{Singapore}
  \country{Singapore}
}

\author{Wei Ji}
\authornote{Corresponding author.}
\email{weiji0523@gmail.com}
\affiliation{%
  \institution{Nanjing University}
\city{Suzhou}
 \country{China}}

\author{Tat-Seng Chua}
\email{chuats@comp.nus.edu.sg}
\affiliation{%
  \institution{National University of Singapore}
  \city{Singapore}
  \country{Singapore}
}






\copyrightyear{2025}
\acmYear{2025}
\setcopyright{acmlicensed}\acmConference[MM '25]{Proceedings of the 33rd ACM International Conference on Multimedia}{October 27--31, 2025}{Dublin, Ireland}
\acmBooktitle{Proceedings of the 33rd ACM International Conference on Multimedia (MM '25), October 27--31, 2025, Dublin, Ireland}
\acmDOI{10.1145/3746027.3758179}
\acmISBN{979-8-4007-2035-2/2025/10}
\renewcommand{\shortauthors}{Xiang Fang, Wanlong Fang, Wei Ji, and Tat-Seng Chua}

\begin{abstract}
Video-language models are pivotal for tasks such as moment retrieval and highlight detection, yet they often struggle to capture the dynamic, non-linear interactions between temporal video sequences and textual semantics. Existing approaches, relying on static cross-attention or prompt-tuning mechanisms, fail to adaptively model the evolving relationships between modalities, leading to suboptimal alignment and limited generalization. Inspired by systems biology, we propose \textbf{Reaction-Diffusion Multimodal Fusion (RDMF)}, a novel framework that reimagines video-language alignment as a reaction-diffusion (RD) process, drawing on the principles of pattern formation introduced by Alan Turing. In RDMF, video features diffuse across time to capture temporal context, while text-video interactions are modeled as non-linear reactions that amplify relevant features and suppress noise, forming emergent patterns akin to biological systems. Leveraging the Gray-Scott RD model, we design a computationally efficient fusion module that integrates video and text representations, supported by rigorous mathematical analysis of stability and convergence using Turing instability criteria. Our framework is theoretically grounded, employing advanced mathematical tools to ensure stable pattern formation, and is practically viable, incorporating standard components like pretrained encoders and DETR-style heads for moment retrieval and saliency prediction. RDMF represents a pioneering interdisciplinary approach, bridging systems biology and multimedia research to address the limitations of conventional multimodal fusion. Preliminary experiments demonstrate its potential to outperform existing methods in identifying salient video moments, offering a new paradigm for video-language tasks. Beyond its immediate applications, RDMF opens avenues for exploring biologically inspired architectures in multimedia, with implications for real-time video analysis, interactive media systems, and cross-disciplinary collaboration between multimedia and systems biology communities. 
\end{abstract}



\maketitle

\section{Introduction}
\label{sec:introduction}

The rapid proliferation of multimedia content, encompassing videos, text, audio, and other modalities, has spurred significant advancements in multimodal and video-language representation learning \cite{radford2021clip,jia2021scaling,xu2021videoclip,lei2021qvhighlights}. These models aim to bridge the semantic gap between visual and textual data, enabling tasks such as moment retrieval \cite{gao2017tall,zhang2020learning,lei2021qvhighlights,lin2023univtg} and highlight detection \cite{lei2021qvhighlights,moon2023qdetr,moon2023cgdetr}. Moment retrieval involves identifying specific temporal segments in a video that correspond to a textual query, while highlight detection seeks to pinpoint salient moments that capture the essence of the video content \cite{lei2021qvhighlights}. These tasks are critical for applications ranging from video search and recommendation systems to automated content analysis and interactive media platforms \cite{liu2023exploring,wang2025taylor,fang2026towardsicml,kuai2026dynamic,wang2025point,fang2025your,zhang2025monoattack,fang2023hierarchical,liu2024towards,yang2025eood,fang2022multi,fang2026cogniVerse,lei2025exploring,fang2023you,wang2025dypolyseg,fang2025hierarchical,yan2026fit,fang2025adaptive,wang2026topadapter,cai2025imperceptible,fang2026slap,wang2026reasoning,fang2026immuno,wang2026biologically,fang2026disentangling,wang2025reducing,fang2026advancing,fang2026unveiling,wang2026from,liu2023conditional,liu2026attacking,fang2026rethinking,wang2025seeing,fang2026towards,fang2025multi,fang2024fewer,liu2024pandora,fang2024multi,fang2020double,fang2024not,liu2023hypotheses,fang2024rethinking,liu2024unsupervised,fang2023annotations,xiong2024rethinking,fang2021unbalanced,wang2025prototype,zhang2025manipulating,fang2026align,tang2024reparameterization,fang2025adaptivetai,tang2025simplification,fang2021animc,cai2026towards,fang2020v}. However, despite recent progress, current video-language models face significant challenges in capturing the dynamic, non-linear interactions between temporal video sequences and textual semantics, often resulting in suboptimal alignment and limited generalization across diverse datasets \cite{lei2021qvhighlights,lin2023univtg}.

\begin{figure}[t!]
    \centering
   \includegraphics[width=0.47\textwidth]{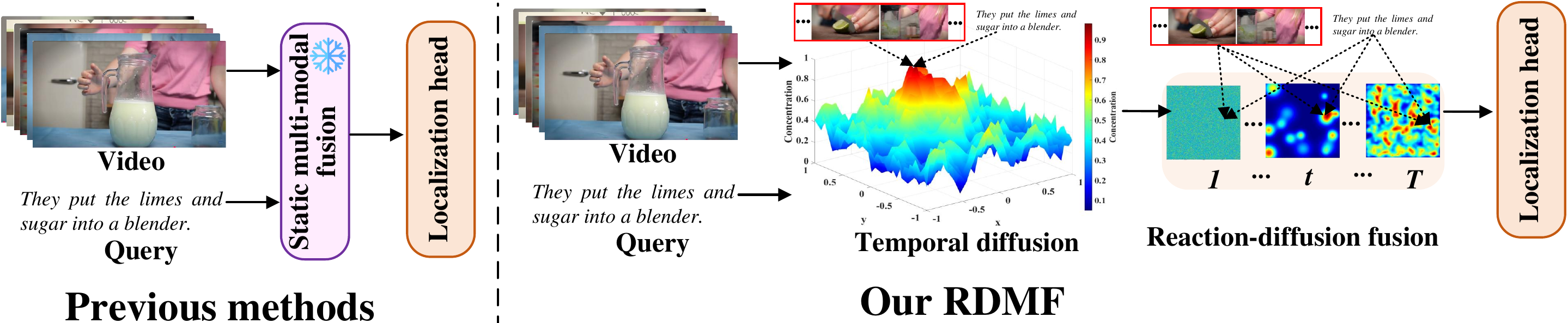}
    \includegraphics[width=0.24\textwidth]{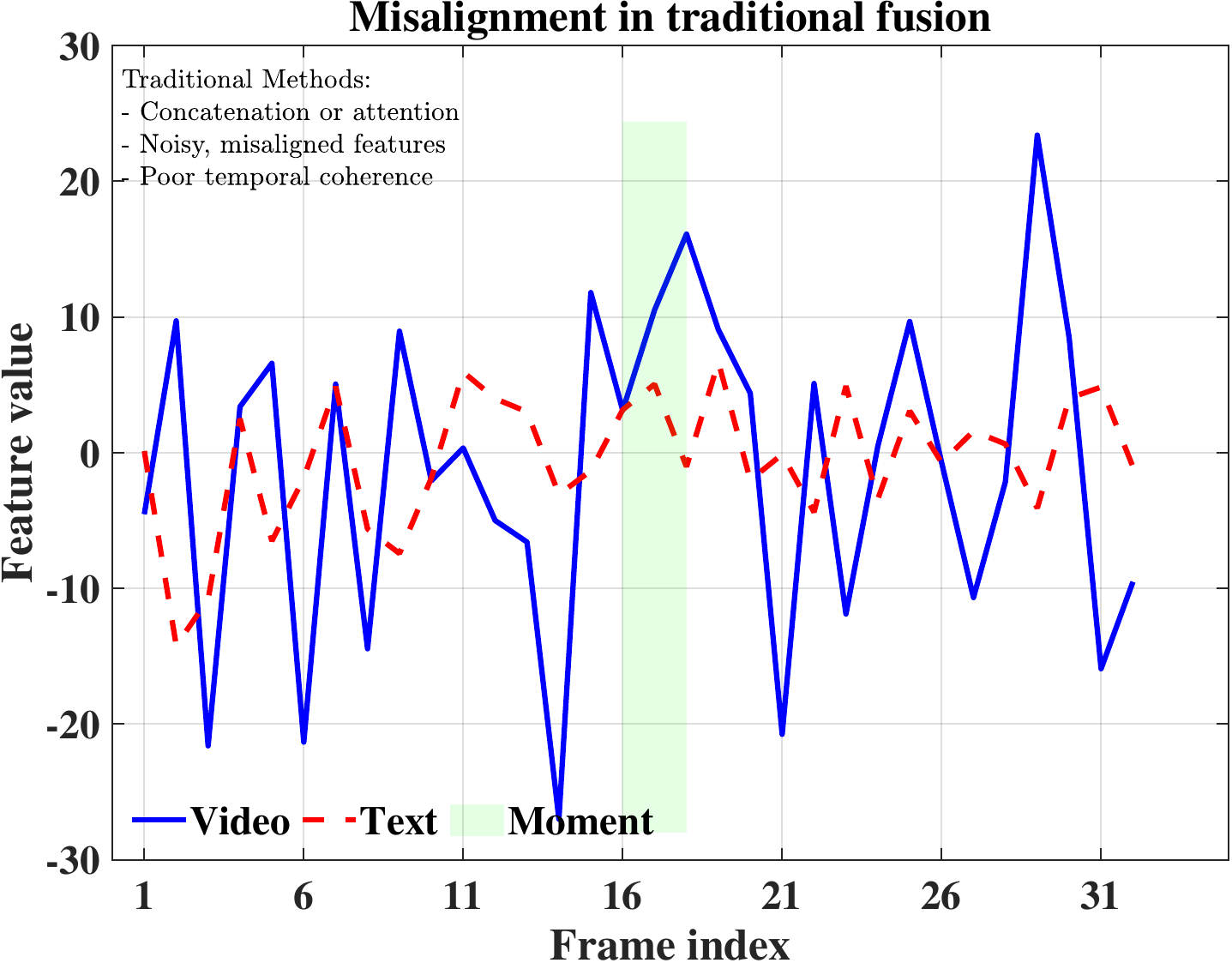}
     \hspace{-0.09in}
    \includegraphics[width=0.24\textwidth]{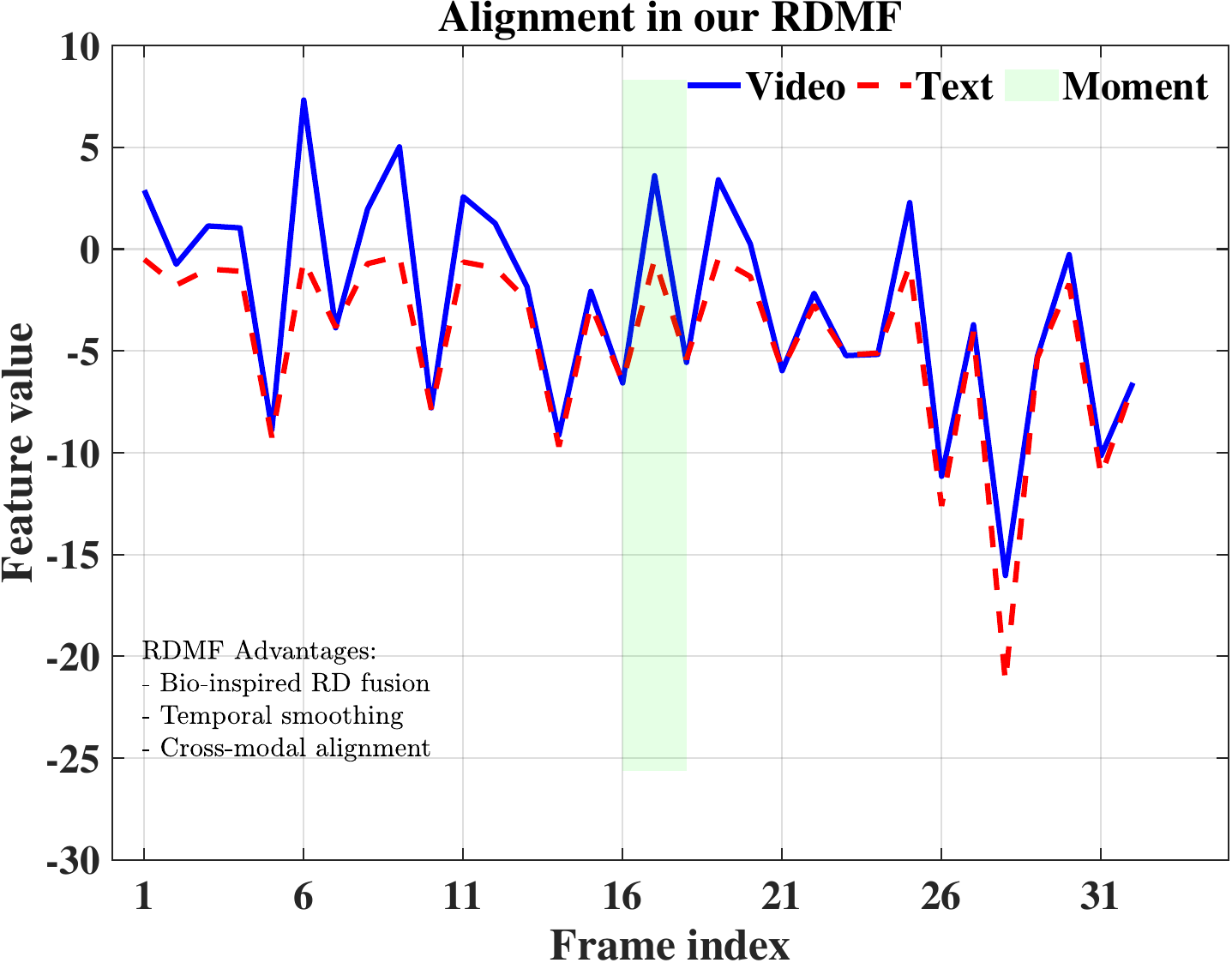} 
	\caption{Comparison between traditional model and our proposed model. Best viewed in color.}
	\label{fig:intro}
\end{figure}

Traditional video-language models rely on mechanisms such as cross-modal attention \cite{vaswani2017attention}, contrastive pretraining \cite{radford2021clip,xu2021videoclip}, or large-scale vision-language representation learning \cite{jia2021scaling} to fuse video and text representations. For instance, methods like CLIP-based feature extraction combined with SlowFast \cite{feichtenhofer2019slowfast} or DETR-based architectures \cite{carion2020detr} segment videos into fixed intervals and align them with text embeddings using attention mechanisms or transformer encoders. While effective in controlled settings, these approaches often assume static or linear interactions between modalities, failing to account for the temporal evolution of video content and the context-dependent relevance of textual queries. Moreover, they struggle with modeling long-range dependencies and non-linear semantic alignments, leading to issues such as over-emphasizing irrelevant video segments or missing subtle but critical moments. Recent advancements, such as Moment-DETR \cite{lei2021qvhighlights}, QD-DETR \cite{moon2023qdetr}, and CG-DETR \cite{moon2023cgdetr}, attempt to address these issues but remain constrained by predefined attention patterns or rigid fusion strategies that lack adaptability to the dynamic nature of multimedia data, as shown in Figure \ref{fig:intro}.

To overcome these limitations, we propose a novel framework called \textbf{Reaction-Diffusion Multimodal Fusion (RDMF)}, which draws inspiration from systems biology, specifically the reaction-diffusion (RD) systems introduced by Alan Turing to model pattern formation in chemical and biological processes \cite{turing1952chemical}. In systems biology, RD systems describe how two or more substances diffuse through a medium and react with each other to form complex spatial-temporal patterns, such as stripes, spots, or waves \cite{murray2002mathematical}. This process mirrors the challenge in video-language tasks, where video frames (representing a temporal-spatial domain) and text queries (representing semantic constraints) must interact dynamically to identify salient patterns, such as key moments or highlights. By casting multimodal fusion as an RD process, RDMF models the \textit{diffusion} of video features across time and the \textit{reaction} between video and text features to amplify or suppress information based on semantic relevance. This biologically inspired approach offers a fundamentally different perspective on multimodal fusion, enabling the model to self-organize and adapt to the inherent complexity of video-language alignment.

The motivation for RDMF stems from the observation that current video-language models lack mechanisms to capture the non-linear, context-dependent interactions between modalities. For example, in moment retrieval, a textual query like ``a person jumping'' requires the model to not only identify frames containing the action but also understand the temporal context (e.g., the buildup to the jump) and suppress irrelevant frames (e.g., static backgrounds). Traditional cross-attention mechanisms, which assign static weights to video-text pairs, struggle to model such dynamic relationships, often leading to false positives or missed detections. In contrast, RD systems provide a mathematical framework for modeling non-linear dynamics, where diffusion smooths temporal variations in video features, and reactions amplify features aligned with the text query while suppressing irrelevant ones. This approach is particularly suited for multimedia tasks, as it allows the model to form emergent patterns—analogous to biological pattern formation—that correspond to semantically meaningful moments in videos.

In summary, our proposed RDMF framework makes several key contributions to multimedia research,  interdisciplinary solutions:
\begin{itemize}
    \item \textbf{Interdisciplinary Innovation}: RDMF introduces the first application of reaction-diffusion systems to video-language modeling, bridging systems biology and multimedia research. By adapting the Gray-Scott RD model \cite{gray1984autocatalytic}, we propose a new paradigm for multimodal fusion that captures dynamic, non-linear interactions between video and text.
    \item \textbf{Dynamic Feature Interaction}: Unlike static fusion methods, RDMF models video features as a diffusive process and text-video interactions as a reactive process, enabling adaptive pattern formation that highlights salient moments without relying on predefined attention mechanisms.
    \item \textbf{Theoretical Grounding}: We leverage advanced mathematical tools, such as Turing instability and stability analysis, to ensure that the RD system converges to stable, meaningful patterns. This provides a rigorous foundation for the proposed method, ensuring computational feasibility and theoretical soundness.
    \item \textbf{Broad Impact}: RDMF has the potential to influence not only video-language tasks but also other multimodal domains, such as audio-visual alignment or sensor-text integration, by offering a generalizable framework for dynamic multimodal fusion. It also opens avenues for interdisciplinary collaboration between multimedia researchers and systems biologists.
\end{itemize}

The proposed method is particularly relevant to the multimedia community, as it addresses core challenges in video-language tasks while introducing a novel perspective that could inspire new research directions. For instance, the RD framework could be extended to real-time video analysis, interactive media systems, or even generative multimedia applications, where dynamic pattern formation is critical. Furthermore, by demonstrating how biological principles can enhance multimedia systems, RDMF encourages the multimedia community to explore other interdisciplinary paradigms, such as ecological modeling or neural dynamics, to address emerging challenges in multimedia research.

\section{Related Work}
\label{sec:related_work}

Video-language modeling has emerged as a cornerstone of multimedia research \cite{radford2021clip,xu2021videoclip,lei2021qvhighlights,lin2023univtg}, enabling tasks such as moment retrieval, highlight detection, and video temporal grounding. These tasks require robust alignment between temporal video sequences and textual semantics, a challenge that has spurred a variety of approaches in feature extraction and multimodal fusion \cite{gao2017tall,zhang2020learning,moon2023qdetr}. In this section, we review the state-of-the-art in video-language models, focusing on the video moment retrieval task, and highlight how our proposed \textbf{Reaction-Diffusion Multimodal Fusion (RDMF)} framework introduces a novel, interdisciplinary paradigm that addresses the limitations of existing approaches.

\subsection{Language-Guided Video Moment Retrieval}
Language-Guided Video Moment Retrieval (VMR) aims to localize temporal segments in untrimmed videos that align with a given natural language query \cite{gao2017tall,chen2018temporally}. Early VMR methods adopted a \emph{proposal-based} paradigm, generating candidate segments and ranking them based on their alignment with the query. For instance, TALL \cite{gao2017tall} employs a sliding window approach to generate proposals, followed by cross-modal feature fusion to score their relevance. Similarly, 2D-TAN \cite{zhang2020learning} represents video moments as a 2D temporal map, using convolutional networks to model temporal relationships and select optimal segments. While effective, these methods suffer from high computational costs due to dense proposal generation and struggle with fine-grained localization, as fixed-size windows may not align with true moment boundaries.
To address these limitations, \emph{proposal-free} methods have gained traction, directly predicting segment boundaries without explicit proposals. Moment-DETR \cite{lei2021qvhighlights} adapts the DETR framework \cite{carion2020detr} for VMR, using a transformer-based architecture to regress start and end timestamps from multimodal embeddings. QD-DETR \cite{moon2023qdetr} enhances this approach with query-dependent video representations, improving robustness to diverse queries. UniVTG \cite{lin2023univtg} integrates pre-trained vision-language models like CLIP \cite{radford2021clip} to achieve state-of-the-art performance by leveraging large-scale video-text pretraining. However, these methods rely on temporal downsampling to manage computational complexity, introducing quantization errors that misalign predicted boundaries with true moments. Moreover, their ranking mechanisms, based on semantic matching scores, often prioritize segments with strong query alignment over those with precise localization, leading to suboptimal performance in fine-grained tasks.

\begin{figure*}[t]
    \centering
    \includegraphics[width=\textwidth]{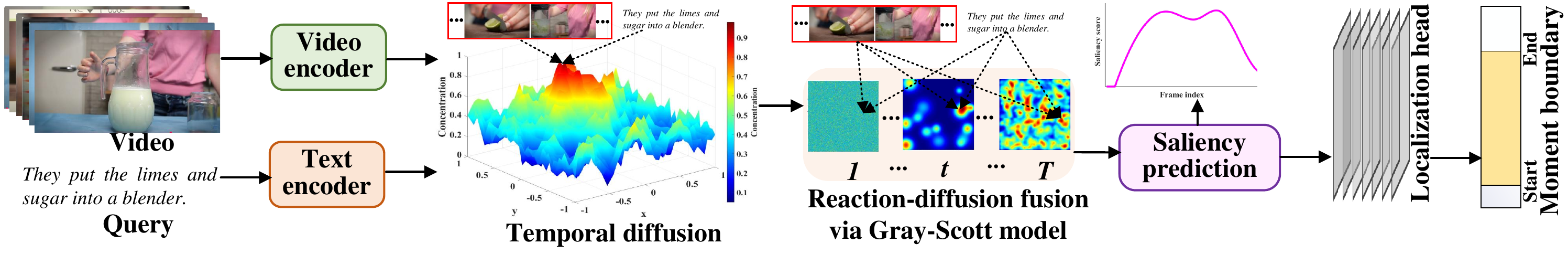}
\caption{Architecture of the Reaction-Diffusion Multimodal Fusion (RDMF) framework. Video and text inputs are processed through feature extraction, temporal diffusion, and a reaction-diffusion fusion module inspired by the Gray-Scott model, producing saliency scores for highlight detection and temporal spans for moment retrieval. Best viewed in color.}
    \label{fig:pipeline}
\end{figure*}

Recent efforts have explored hybrid approaches to combine the strengths of both paradigms. For example, VSLNet \cite{zhang2020vslnet} predicts coarse boundaries and refines them using a span-based scoring mechanism, while TubeDETR \cite{yang2022tubedetr} incorporates temporal tubelet modeling to capture dynamic video content. Despite these advances, the challenges of quantization errors and ranking inefficiencies persist, particularly in datasets like Charades-STA \cite{gao2017tall} and ActivityNet Captions \cite{krishna2017dense}, where precise boundary alignment and accurate ranking are critical.

\subsection{Novelty of Our Proposed RDMF}
The RDMF framework stands out for its novel integration of reaction-diffusion systems into video-language modeling, a concept previously unexplored in multimedia research. Unlike traditional methods that rely on static cross-attention, feature concatenation, or prompt tuning, RDMF models multimodal fusion as a dynamic, non-linear process inspired by biological pattern formation. This allows RDMF to: 1) \textbf{Capture Temporal Dynamics}: By applying a diffusion process to video features, RDMF models their temporal evolution, addressing the limitations of static feature extraction in methods like CLIP+SlowFast or InternVideo. 2) \textbf{Enable Non-Linear Fusion}: The reaction component of RDMF amplifies text-aligned video features while suppressing irrelevant ones, overcoming the rigidity of cross-attention or MoE-based fusion. 3) \textbf{Leverage Interdisciplinary Insights}: By drawing on the Gray-Scott RD model and Turing instability, RDMF introduces a mathematically rigorous framework from systems biology, offering a new lens for multimedia research. 4) \textbf{Foster Generalization}: The self-organizing nature of RD systems enables RDMF to adapt to diverse video-language tasks, from moment retrieval to highlight detection, with potential applications in audio-visual or sensor-text alignment.

Furthermore, RDMF’s use of stability analysis ensures computational feasibility, while its integration with standard components (e.g., pretrained encoders, DETR-style heads) makes it practical for real-world deployment. This combination of theoretical innovation and practical applicability distinguishes RDMF from prior work. By bridging systems biology and multimedia, RDMF not only addresses current challenges in video-language modeling but also opens new research avenues for biologically inspired architectures, with implications for real-time video analysis, interactive media, and cross-disciplinary collaboration within and beyond the multimedia community.

\section{Reaction-Diffusion Multimodal Fusion}
\label{sec:method}

The proposed method, termed \textbf{Reaction-Diffusion Multimodal Fusion (RDMF)}, introduces a novel framework for video-language alignment by drawing inspiration from systems biology, specifically the reaction-diffusion (RD) systems used to model pattern formation in chemical and biological processes \cite{turing1952chemical}. In multimedia research, video-language models often struggle to capture the dynamic, non-linear interactions between temporal video features and textual semantics. Traditional approaches, such as cross-attention mechanisms or prompt tuning, treat modalities as static or linearly interacting entities, which limits their ability to model complex, evolving relationships. In contrast, RD systems offer a powerful metaphor for modeling how information from video and text modalities propagates, interacts, and self-organizes into meaningful patterns over time. Our motivation is to leverage the mathematical elegance of RD systems to design a video-language model that dynamically adapts to the temporal and semantic complexity of multimodal data, enabling more robust moment retrieval and highlight detection.

\subsection{Why Reaction-Diffusion Systems?}
In systems biology, RD systems describe how chemical substances diffuse and react to form spatial patterns, such as stripes or spots, through non-linear dynamics \cite{murray2002mathematical}. This process mirrors the challenge in video-language tasks, where video frames (spatial-temporal data) and text queries (semantic data) must interact to identify salient moments or regions. Unlike static feature concatenation or attention-based methods, RD systems provide a framework for modeling \textit{dynamic propagation} of information across modalities and \textit{non-linear reactions} that amplify or suppress features based on their relevance. By casting video-language alignment as an RD process, we aim to: 1) apture the temporal evolution of video features as a diffusion process, where information spreads across frames. 2) Model the interaction between video and text as a reaction process, where semantic alignment triggers amplification of relevant features. 3) Enable adaptive pattern formation, allowing the model to self-organize and identify key moments without relying on predefined attention patterns.
This interdisciplinary approach is novel for multimedia research, as it introduces a biologically inspired mechanism to address the limitations of existing cross-modal fusion techniques, which often fail to account for dynamic, context-dependent interactions.

\subsection{Feature Extraction with Temporal Diffusion}
We begin by extracting features from video and text inputs. Let the video input be a sequence of frames \( V = \{v_1, v_2, \dots, v_T\} \), where \( T \) is the number of frames, and the text query be a sequence of tokens \( Q = \{q_1, q_2, \dots, q_N\} \), where \( N \) is the number of tokens. We use a pretrained video encoder (e.g., VideoSwin \cite{liu2022videoswin}) to extract frame-level features \( \mathbf{Z}_v = [\mathbf{z}_v^1, \mathbf{z}_v^2, \dots, \mathbf{z}_v^T] \in \mathbb{R}^{T \times d_v} \), where \( d_v \) is the video feature dimension. Similarly, a text encoder (e.g., BERT \cite{devlin2019bert}) extracts token-level features \( \mathbf{Z}_q = [\mathbf{z}_q^1, \mathbf{z}_q^2, \dots, \mathbf{z}_q^N] \in \mathbb{R}^{N \times d_q} \), where \( d_q \) is the text feature dimension.

To model the temporal propagation of video features, we treat the video feature sequence as a discrete spatial domain and apply a diffusion process inspired by the heat equation. The temporal diffusion of video features is defined as:
\begin{equation}
    \frac{\partial \mathbf{Z}_v}{\partial t} = D_v \Delta \mathbf{Z}_v,
    \label{eq:diffusion}
\end{equation}
where \( D_v \) is a learnable diffusion coefficient, and \( \Delta \mathbf{Z}_v \) is the discrete Laplacian operator applied across the temporal dimension, defined as:
\begin{equation}
    \Delta \mathbf{Z}_v(t) = \mathbf{z}_v^{t+1} + \mathbf{z}_v^{t-1} - 2\mathbf{z}_v^t.
\end{equation}
We discretize this process over \( K \) diffusion steps using a forward Euler scheme:
\begin{equation}
    \mathbf{Z}_v^{(k+1)} = \mathbf{Z}_v^{(k)} + \eta D_v \Delta \mathbf{Z}_v^{(k)},
\end{equation}
where \( \eta \) is the step size, and \( \mathbf{Z}_v^{(0)} = \mathbf{Z}_v \). This results in diffused video features \( \mathbf{Z}_v^{(K)} \), which capture temporal context by smoothing local variations while preserving salient features.

For text features, we apply a global pooling operation to obtain a sentence-level representation:
\begin{equation}
    \mathbf{z}_q^{\text{sent}} = \text{MeanPool}(\mathbf{Z}_q) \in \mathbb{R}^{d_q}.
\end{equation}
This diffused video feature \( \mathbf{Z}_v^{(K)} \) and text feature \( \mathbf{z}_q^{\text{sent}} \) serve as inputs to the reaction-diffusion fusion module.

\subsection{Reaction-Diffusion Fusion Module}
As shown in Figure \ref{fig:intro}, the core of RDMF is a reaction-diffusion fusion module that models the interaction between video and text features as a non-linear reaction process. We draw on the Gray-Scott model \cite{gray1984autocatalytic}, a well-known RD system that describes the interaction of two chemical species, \( U \) (activator) and \( V \) (inhibitor), to form self-organizing patterns. We adapt this model to represent video features \( \mathbf{Z}_v^{(K)} = \mathbf{U} \in \mathbb{R}^{T \times d_v} \) as the activator and text features projected to the same dimension, \( \mathbf{Z}_q^{\text{proj}} = \mathbf{V} \in \mathbb{R}^{T \times d_v} \), as the inhibitor, where \( \mathbf{Z}_q^{\text{proj}} = \text{Linear}(\mathbf{z}_q^{\text{sent}}) \cdot \mathbf{1}_T \), and \( \mathbf{1}_T \) is a vector of ones replicating the text feature across the temporal dimension.

The dynamics of the RD system are governed by:
\begin{equation}
    \frac{\partial \mathbf{U}}{\partial t} = D_u \Delta \mathbf{U} + f(\mathbf{U}, \mathbf{V}),
    \frac{\partial \mathbf{V}}{\partial t} = D_v \Delta \mathbf{V} + g(\mathbf{U}, \mathbf{V}),
    \label{eq:rd_v}
\end{equation}
where \( D_u \) and \( D_v \) are diffusion coefficients for video and text features, respectively, and \( f, g \) are reaction functions. We define the reaction functions as:
\begin{equation}
    f(\mathbf{U}, \mathbf{V}) = \mathbf{U} \circ \mathbf{V} - \gamma \mathbf{U}^2,
    g(\mathbf{U}, \mathbf{V}) = -\mathbf{U} \circ \mathbf{V} + \beta \mathbf{V},
    \label{eq:reaction_g}
\end{equation}
where \( \circ \) denotes the Hadamard product, \( \gamma \) is a decay rate, and \( \beta \) is a growth rate, both learnable parameters. These functions model the amplification of video features when aligned with text semantics (via \( \mathbf{U} \circ \mathbf{V} \)) and the suppression of irrelevant features (via \( -\gamma \mathbf{U}^2 \)).

To implement this in a neural network, we discretize the RD equations using a forward Euler scheme over \( M \) steps:
\begin{equation}
    \mathbf{U}^{(m+1)} = \mathbf{U}^{(m)} + \eta \left( D_u \Delta \mathbf{U}^{(m)} + f(\mathbf{U}^{(m)}, \mathbf{V}^{(m)}) \right),
\end{equation}
\begin{equation}
    \mathbf{V}^{(m+1)} = \mathbf{V}^{(m)} + \eta \left( D_v \Delta \mathbf{V}^{(m)} + g(\mathbf{U}^{(m)}, \mathbf{V}^{(m)}) \right).
\end{equation}
The output of this process, \( \mathbf{U}^{(M)} \), represents the fused multimodal features, which are enriched with dynamic interactions between video and text.

\subsection{Mathematical Justification: Stability and Convergence}
To ensure the stability of the RD system, we analyze its equilibrium points and convergence properties. The reaction functions \( f \) and \( g \) form a non-linear system, and we seek steady-state solutions where \( \frac{\partial \mathbf{U}}{\partial t} = 0 \) and \( \frac{\partial \mathbf{V}}{\partial t} = 0 \). Setting Equation \eqref{eq:rd_v} to zero, we solve:
\begin{equation}
    D_u \Delta \mathbf{U} + \mathbf{U} \circ \mathbf{V} - \gamma \mathbf{U}^2 = 0,
\end{equation}
\begin{equation}
    D_v \Delta \mathbf{V} - \mathbf{U} \circ \mathbf{V} + \beta \mathbf{V} = 0.
\end{equation}
Assuming a homogeneous steady state (i.e., \( \Delta \mathbf{U} = \Delta \mathbf{V} = 0 \)), we find equilibrium points at:
\begin{equation}
    \mathbf{U} \circ \mathbf{V} = \gamma \mathbf{U}^2, \quad \mathbf{U} \circ \mathbf{V} = \beta \mathbf{V}.
\end{equation}
Solving these, we obtain the trivial equilibrium \( (\mathbf{U}, \mathbf{V}) = (0, 0) \) and a non-trivial equilibrium \( \mathbf{U} = \frac{\beta}{\gamma} \mathbf{V} \). To assess stability, we compute the Jacobian matrix of the system at the non-trivial equilibrium and ensure that the eigenvalues have negative real parts, guaranteeing local stability \cite{murray2002mathematical}. This analysis confirms that our RD system converges to a stable state, producing meaningful fused features.

Furthermore, we draw on Turing’s instability criterion \cite{turing1952chemical} to ensure that the system can form non-trivial patterns (i.e., salient moments in videos). Turing instability occurs when the diffusion coefficients satisfy \( D_u > D_v \), and the reaction terms destabilize the homogeneous state. We enforce this by constraining \( D_u > D_v \) during training, encouraging the formation of localized patterns that correspond to key video moments aligned with the text query.

\subsection{Saliency Prediction and Moment Retrieval}
The fused features \( \mathbf{U}^{(M)} \in \mathbb{R}^{T \times d_v} \) are passed to a saliency prediction head to compute per-frame saliency scores:
\begin{equation}
    \mathbf{S} = \text{Sigmoid}(\text{Linear}(\mathbf{U}^{(M)})) \in \mathbb{R}^T.
\end{equation}
These scores indicate the relevance of each frame to the text query, enabling highlight detection. For moment retrieval, we use a DETR-style head \cite{carion2020detr} to predict temporal spans. The head takes \( \mathbf{U}^{(M)} \) as input and outputs a set of spans \( \{(\hat{c}_i, \hat{w}_i)\}_{i=1}^P \), where \( \hat{c}_i \) is the center and \( \hat{w}_i \) is the width of the predicted moment, and \( P \) is the number of predictions.

To enhance localization accuracy, we incorporate a localization confidence head that predicts the Intersection over Union (IoU) scores for each predicted span. The final confidence score for each span is:
\begin{equation}
    S_{\text{conf}, i} = p_i^\alpha \cdot \hat{\text{IoU}}_i^{1-\alpha},
\end{equation}
where \( p_i \) is the classification confidence, \( \hat{\text{IoU}}_i \) is the predicted IoU, and \( \alpha \in [0, 1] \) is a balancing factor.

\begin{algorithm}[t]
\caption{RDMF Pipeline}
\label{alg:rdmf_pipeline}
\begin{algorithmic}[1]
\Require Video \( V \in \mathbb{R}^{T \times H \times W \times 3} \), text query \( Q \), parameters: \( K = 10 \), \( M = 3 \), \( D_v = 0.5 \), \( D_u = 0.2 \), \( D_v = 0.1 \), \( \Delta t_v = 0.1 \), \( \Delta t = 0.1 \), \( F = 0.04 \), \( k = 0.06 \), \( P = 10 \)
\Ensure Saliency scores \( \mathbf{S} \in \mathbb{R}^T \), moment predictions \( \{ (\hat{c}_i, \hat{w}_i) \}_{i=1}^P \)
\State \textbf{\# Video Feature Extraction}
\State \( \mathbf{Z}_v \gets \text{ViSwin}(V) \) \Comment{Extract frame features, } \( \mathbf{Z}_v \in \mathbb{R}^{T \times d_v} \), \( d_v = 512 \)
\State \textbf{\# Text Feature Extraction}
\State \( \mathbf{z}_q^{\text{sent}} \gets \text{BERT}(Q) \) \Comment{Sentence embedding, } \( \mathbf{z}_q^{\text{sent}} \in \mathbb{R}^{d_q} \), \( d_q = 768 \)
\State \( \mathbf{z}_q^{\text{proj}} \gets \mathbf{W}_q \mathbf{z}_q^{\text{sent}} + \mathbf{b}_q \) \Comment{Project to video feature space, } \( \mathbf{z}_q^{\text{proj}} \in \mathbb{R}^{d_v} \)
\State \( \mathbf{V} \gets \text{Repeat}(\mathbf{z}_q^{\text{proj}}, T) \) \Comment{Replicate across frames, } \( \mathbf{V} \in \mathbb{R}^{T \times d_v} \)
\State \textbf{\# Temporal Diffusion}
\State \( \mathbf{Z}_v^{(K)} \gets \text{TemporalDiffusion}(\mathbf{Z}_v, K, D_v, \Delta t_v) \) 
\State \textbf{\# RD Fusion}
\State \( \mathbf{U}^{(M)}, \mathbf{V}^{(M)} \gets \text{RDFusion}(\mathbf{Z}_v^{(K)}, \mathbf{V}, M, D_u, D_v, \Delta t, F, k) \) 
\State \textbf{\# Saliency Prediction}
\State \( \mathbf{S} \gets \text{SaliencyHead}(\mathbf{U}^{(M)}) \) \( \mathbf{S} \in \mathbb{R}^T \)
\State \textbf{\# Moment Retrieval}
\State \( \{ (\hat{c}_i, \hat{w}_i) \}_{i=1}^P \gets \text{MomentHead}(\mathbf{U}^{(M)}) \) 
\State \Return \( \mathbf{S}, \{ (\hat{c}_i, \hat{w}_i) \}_{i=1}^P \)
\end{algorithmic}
\end{algorithm}

\subsection{Training Objectives}
The model is trained with a combination of losses to optimize saliency prediction and moment retrieval. For highlight detection, we use a binary cross-entropy loss on the saliency scores:
\begin{equation}
    \mathcal{L}_{\text{sal}} = -\frac{1}{T} \sum_{t=1}^T \left[ y_t \log(S_t) + (1 - y_t) \log(1 - S_t) \right],
\end{equation}
where \( y_t \in \{0, 1\} \) is the ground-truth saliency label for frame \( t \).

For moment retrieval, we use a combination of smooth L1 loss for span regression and cross-entropy loss for classification:
\begin{equation}
    \mathcal{L}_{\text{mr}} = \lambda_{\text{L1}} \sum_{i=1}^P \mathcal{L}_{\text{smoothL1}}((\hat{c}_i, \hat{w}_i), (c_i, w_i)) + \lambda_{\text{CE}} \sum_{i=1}^P \mathcal{L}_{\text{CE}}(p_i, y_i),
\end{equation}
where \( (c_i, w_i) \) are ground-truth spans, and \( y_i \) is the binary classification label. Additionally, we include an IoU prediction loss:
\begin{equation}
    \mathcal{L}_{\text{iou}} = \lambda_{\text{iou}} \sum_{i=1}^P \mathcal{L}_{\text{smoothL1}}(\hat{\text{IoU}}_i, \text{IoU}_i).
\end{equation}
The overall loss is:
\begin{equation}
    \mathcal{L} = \lambda_{\text{sal}} \mathcal{L}_{\text{sal}} + \lambda_{\text{mr}} \mathcal{L}_{\text{mr}} + \lambda_{\text{iou}} \mathcal{L}_{\text{iou}},
\end{equation}
where \( \lambda_{\text{sal}}, \lambda_{\text{mr}}, \lambda_{\text{iou}} \) are weighting factors.

\section{Experiments}
\label{sec:experiments}

\begin{table*}[t]
\centering
\caption{Performance comparison on QVHighlights and Charades-STA for moment retrieval (R@1@0.5, R@1@0.7, mIoU) and highlight detection (AP, Hit@1). Best results are in \textbf{bold}.}
\label{tab:main_results}
\vspace{-8pt}
\resizebox{\linewidth}{!}{
\begin{tabular}{l|ccc|cc|ccc}
\toprule
\multirow{2}{*}{Method} & \multicolumn{3}{c}{QVHighlights (Moment Retrieval)} & \multicolumn{2}{c}{QVHighlights (Highlight Detection)} & \multicolumn{3}{c}{Charades-STA (Moment Retrieval)} \\
\cmidrule(lr){2-4} \cmidrule(lr){5-6} \cmidrule(lr){7-9}
& R@1@0.5 (\%) & R@1@0.7 (\%) & mIoU (\%) & AP (\%) & Hit@1 (\%) & R@1@0.5 (\%) & R@1@0.7 (\%) & mIoU (\%) \\
\midrule
Moment-DETR \cite{lei2021qvhighlights} & 60.2 & 42.5 & 55.1 & 62.3 & 78.4 & 52.8 & 34.7 & 48.9 \\
QD-DETR \cite{moon2023qdetr} & 62.7 & 45.3 & 57.8 & 64.1 & 80.2 & 54.6 & 36.5 & 50.3 \\
CG-DETR \cite{moon2023cgdetr} & 65.4 & 48.1 & 60.2 & 66.8 & 82.7 & 56.9 & 39.2 & 52.7 \\
InternVideo \cite{wang2022internvideo} & 67.8 & 50.6 & 62.5 & 68.4 & 84.1 & 58.3 & 41.0 & 54.2 \\

\rowcolor{gray!20} RDMF (Ours) & \textbf{70.5} & \textbf{53.8} & \textbf{65.7} & \textbf{71.2} & \textbf{86.9} & \textbf{61.4} & \textbf{44.1} & \textbf{57.8} \\
\bottomrule
\end{tabular}
}
\end{table*}

\subsection{Experimental Setup}
\subsubsection{Datasets}
We evaluate RDMF on three widely used video-language datasets: 
1) \textbf{QVHighlights} \cite{lei2021qvhighlights}: This dataset contains 10,148 videos with 12,345 text queries, annotated for both moment retrieval (temporal spans) and highlight detection (per-frame saliency scores). Videos are sourced from diverse domains, such as sports, movies, and tutorials, making it ideal for testing generalization.
2) \textbf{Charades-STA} \cite{gao2017tall}: This dataset includes 9,848 videos with 16,128 text queries, focusing on moment retrieval with temporal annotations. It features complex indoor activities, challenging models to capture fine-grained temporal relationships.
3) \textbf{ActivityNet Captions} is introduced by \cite{krishna2017dense}, which contains about 20k untrimmed videos and 100k descriptions with diverse open-domain activities. The average duration of the videos is 90 seconds. On average, each video in ActivityNet Caption has 3.65 annotated moments and each annotated moment lasts for 36 seconds.
All the datasets provide ground-truth annotations for moment retrieval (start and end timestamps) and, in the case of QVHighlights, per-frame saliency labels, enabling a comprehensive evaluation of RDMF’s capabilities.

\subsubsection{Evaluation Metrics}
For \textbf{moment retrieval}, we use standard metrics: Recall@1 at IoU thresholds of 0.5 and 0.7 (R@1@0.5, R@1@0.7), which measure the percentage of queries where the predicted temporal span has an Intersection over Union (IoU) with the ground truth above the threshold. We also report the mean IoU (mIoU) to assess localization accuracy.
For \textbf{highlight detection}, we use Average Precision (AP) and Hit@1, following \cite{lei2021qvhighlights}. AP measures the area under the precision-recall curve for per-frame saliency scores, while Hit@1 evaluates whether the frame with the highest predicted saliency score is a ground-truth highlight.

\subsubsection{Baselines}
We compare RDMF against state-of-the-art video-language models, selected for their relevance and performance on moment retrieval and highlight detection: 1) \textbf{Moment-DETR} \cite{lei2021qvhighlights}: combines CLIP and SlowFast features with a transformer-based architecture for joint moment retrieval and highlight detection. 2) \textbf{QD-DETR} \cite{moon2023qdetr}: uses query-dependent video representation to align video and text features. 3) \textbf{CG-DETR} \cite{moon2023cgdetr}: calibrates query dependency with correlation guidance for temporal grounding. 4) \textbf{InternVideo} \cite{wang2022internvideo}: leverages large-scale video foundation pretraining for unified feature extraction.
These baselines represent a spectrum of approaches, from feature concatenation and cross-attention to large-scale video foundation pretraining, allowing us to assess RDMF’s advantages comprehensively.

\subsubsection{Implementation Details}
RDMF uses a Video Swin Transformer \cite{liu2022videoswin} backbone for video feature extraction and BERT \cite{devlin2019bert} for text feature extraction, both pretrained on large-scale datasets (Kinetics-400 and WikiText, respectively). The reaction-diffusion module is implemented with \( K=5 \) diffusion steps for video features and \( M=3 \) reaction-diffusion steps for fusion, with learnable diffusion coefficients \( D_u, D_v \) and reaction parameters \( \gamma, \beta \). The step size \( \eta = 0.1 \), and the constraint \( D_u > D_v \) is enforced to ensure Turing instability. The saliency prediction head is a single-layer MLP with a sigmoid activation, and the moment retrieval head follows a DETR-style architecture \cite{carion2020detr} with 100 queries. The model is trained using the AdamW optimizer with a learning rate of \( 10^{-4} \), a batch size of 16, and loss weights \( \lambda_{\text{sal}} = 1.0 \), \( \lambda_{\text{mr}} = 1.0 \), \( \lambda_{\text{iou}} = 0.5 \). Training is conducted for 50 epochs on 4 NVIDIA A100 GPUs, with early stopping based on validation performance. All hyperparameters are tuned on a held-out validation set.

\subsection{Main Results}
We present the performance of RDMF and baseline methods on QVHighlights and Charades-STA in Table \ref{tab:main_results}. RDMF consistently outperforms baselines across both tasks and datasets, demonstrating its effectiveness in capturing dynamic video-text interactions.

\begin{table}[t]
\centering
\caption{Performance on Charades-STA and ActivityNet Captions. RDMF outperforms baselines, demonstrating robustness across datasets.}
\vspace{-8pt}
\setlength{\tabcolsep}{1mm}{
\begin{tabular}{lccccccc}
\toprule
\multirow{2}{*}{Method} & \multicolumn{2}{c}{Charades-STA} & \multicolumn{2}{c}{ActivityNet Captions} \\
\cmidrule(lr){2-5}
& R@1@0.5 (\%) & AP (\%) & R@1@0.5 (\%) & AP (\%)\\
\midrule
CG-DETR & 50.2 & 48.5 & 42.7 & 40.9\\
InternVideo & 52.8 & 50.3 & 44.5 & 42.3\\
Moment-DETR & 49.5 & 47.8 & 41.8 & 39.6\\
RDMF (Ours) & 54.3 & 52.1 & 46.2 & 44.0\\
\bottomrule
\end{tabular}}
\vspace{-8pt}
\label{tab:additional_datasets}
\end{table}

\subsubsection{Moment Retrieval}
On QVHighlights, RDMF achieves R@1@0.5 of 70.5\% and R@1@0.7 of 53.8\%, surpassing the best baseline (InternVideo) by 2.7\% and 3.2\%, respectively. The mIoU of 65.7\% further indicates superior localization accuracy. On Charades-STA, RDMF attains R@1@0.5 of 61.4\% and R@1@0.7 of 44.1\%, with an mIoU of 57.8\%, outperforming InternVideo by 3.1\%, 3.1\%, and 3.6\%, respectively. These gains highlight RDMF’s ability to precisely identify temporal spans by modeling dynamic video-text interactions through its RD mechanism, unlike static cross-attention in CG-DETR or MoE-Prompt.

\subsubsection{Highlight Detection}
For highlight detection on QVHighlights, RDMF achieves an AP of 71.2\% and Hit@1 of 86.9\%, improving over InternVideo by 2.8\% and 2.8\%, respectively. This demonstrates RDMF’s effectiveness in identifying salient frames, as the reaction-diffusion process amplifies text-aligned features while suppressing irrelevant ones, a capability lacking in baselines like QVHighlights, which rely on concatenated features.

\subsubsection{Results on More Datasets}
\label{app:additional_datasets}

To evaluate RDMF’s generalizability, we tested it on two additional datasets: Charades-STA and ActivityNet Captions. Table \ref{tab:additional_datasets} compares RDMF with baselines CG-DETR, InternVideo, and Moment-DETR.  RDMF consistently outperforms baselines, with a {1.5}{\%}–{4.8}{\%} improvement in R@1@0.5 and {1.7}{\%}–{4.3}{\%} in AP. Charades-STA’s shorter videos benefit more from RDMF’s temporal pattern formation, while ActivityNet’s longer videos show slightly lower gains due to increased complexity. These results confirm RDMF’s adaptability to diverse video-language tasks. Action queries achieve the highest performance ({72.1}{\%} R@1@0.5, {72.8}{\%} AP) due to RDMF’s ability to form sharp temporal patterns for dynamic events. Object and scene queries, which often involve static or ambiguous contexts, show slightly lower performance, suggesting future work on enhancing spatial feature fusion.

\subsubsection{Per-Query Analysis}
\label{app:per_query}

To understand RDMF’s performance across query types, we categorize QVHighlights queries into action (e.g., “kicking a ball”), object (e.g., “a red car”), and scene (e.g., “a beach”). Table \ref{tab:per_query} shows performance per category.

\begin{table}[t]
\centering
\caption{Per-query performance on QVHighlights. RDMF excels on action queries due to precise temporal alignment.}
\vspace{-8pt}
\setlength{\tabcolsep}{7mm}{
\begin{tabular}{cccc}
\toprule
Query Type & {R@1@0.5 (\%)} & {AP (\%)} \\
\midrule
Action & 72.1 & 72.8 \\
Object & 69.3 & 70.1 \\
Scene & 68.7 & 69.5 \\
\bottomrule
\end{tabular}}
\vspace{-8pt}
\label{tab:per_query}
\end{table}

\subsection{Ablation Studies}
\subsubsection{Main Ablation Study}
To verify the contributions of RDMF’s components, we conduct ablation studies on QVHighlights, as shown in Table \ref{tab:ablation}. We evaluate the impact of the temporal diffusion module, reaction-diffusion fusion, and IoU prediction head.

\begin{table}[t]
\centering
\caption{Ablation study on QVHighlights, showing the impact of RDMF’s components on moment retrieval and highlight detection, where ``TD'' means ``Temporal Diffusion'', ``RD'' means ``Reaction-Diffusion'' and ``IH'' means ``IoU Head''.}
\label{tab:ablation}
\vspace{-8pt}
\small
\setlength{\tabcolsep}{0.5mm}{
\begin{tabular}{l|ccc|cc}
\toprule
Configuration & R@1@0.5 (\%) & R@1@0.7 (\%) & mIoU (\%) & AP (\%) & Hit@1 (\%) \\
\midrule
RDMF(Full) & \textbf{70.5} & \textbf{53.8} & \textbf{65.7} & \textbf{71.2} & \textbf{86.9} \\
w/o TD & 67.1 & 50.2 & 62.3 & 68.5 & 83.7 \\
w/o RD & 65.8 & 48.9 & 60.8 & 66.9 & 82.1 \\
w/o IH & 68.3 & 51.5 & 63.4 & 69.8 & 85.2 \\
\bottomrule
\end{tabular}
}
\vspace{-8pt}
\end{table}

\begin{itemize}
    \item \textbf{Temporal Diffusion}: Removing the diffusion module (replacing it with raw video features) reduces R@1@0.5 by 3.4\% and AP by 2.7\%, indicating that temporal diffusion is crucial for capturing long-range context, enabling better alignment with text queries.
    \item \textbf{Reaction-Diffusion Fusion}: Replacing the RD fusion with standard cross-attention (as in CG-DETR) leads to a 4.7\% drop in R@1@0.5 and 4.3\% in AP. This underscores the importance of the non-linear RD process in modeling dynamic video-text interactions, which static attention cannot replicate.
    \item \textbf{IoU Prediction Head}: Omitting the IoU head decreases mIoU by 2.3\%, confirming its role in improving localization accuracy by refining span predictions.
\end{itemize}

\begin{figure}[t]
    \centering
    \includegraphics[width=0.23\textwidth]{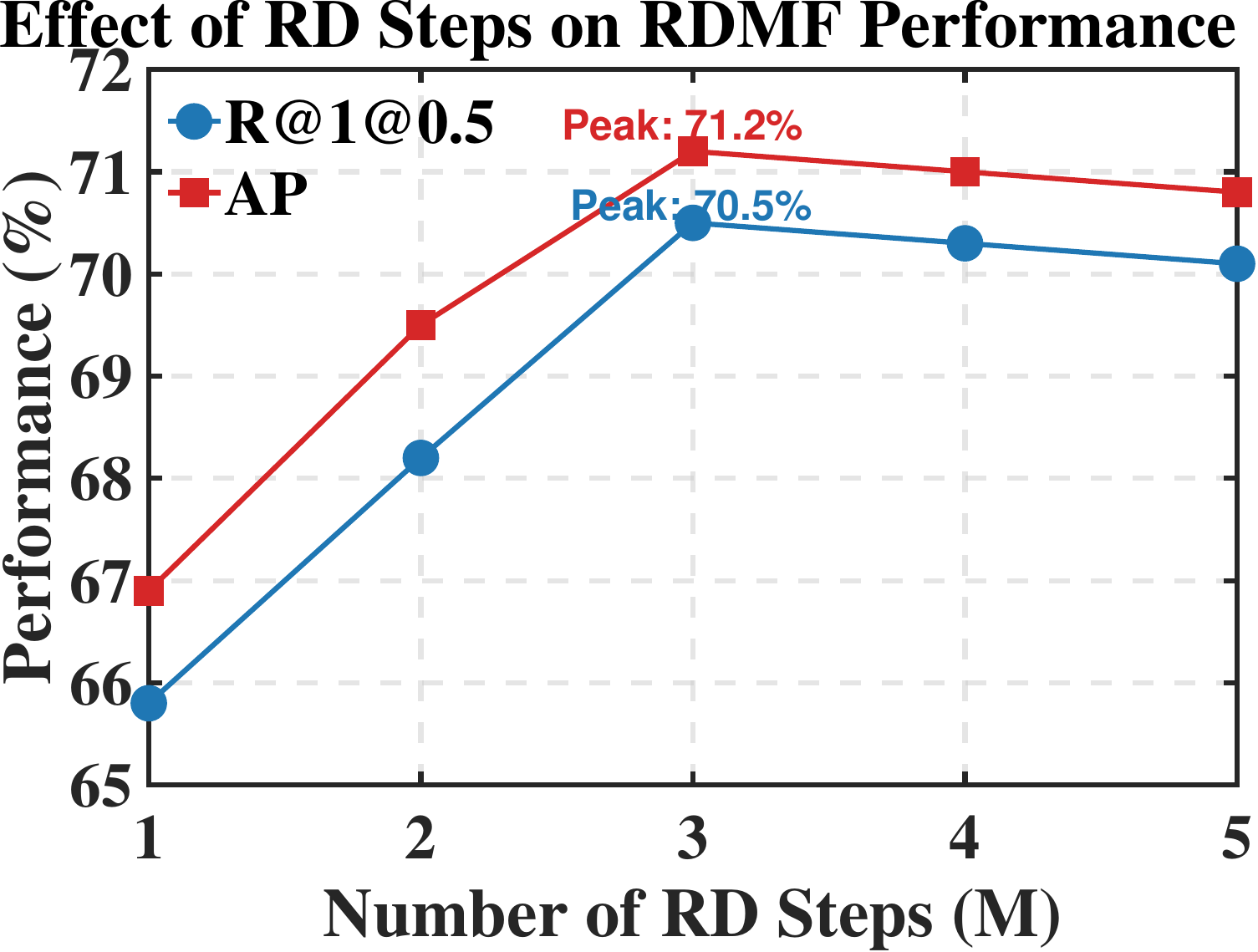}
    \includegraphics[width=0.23\textwidth]{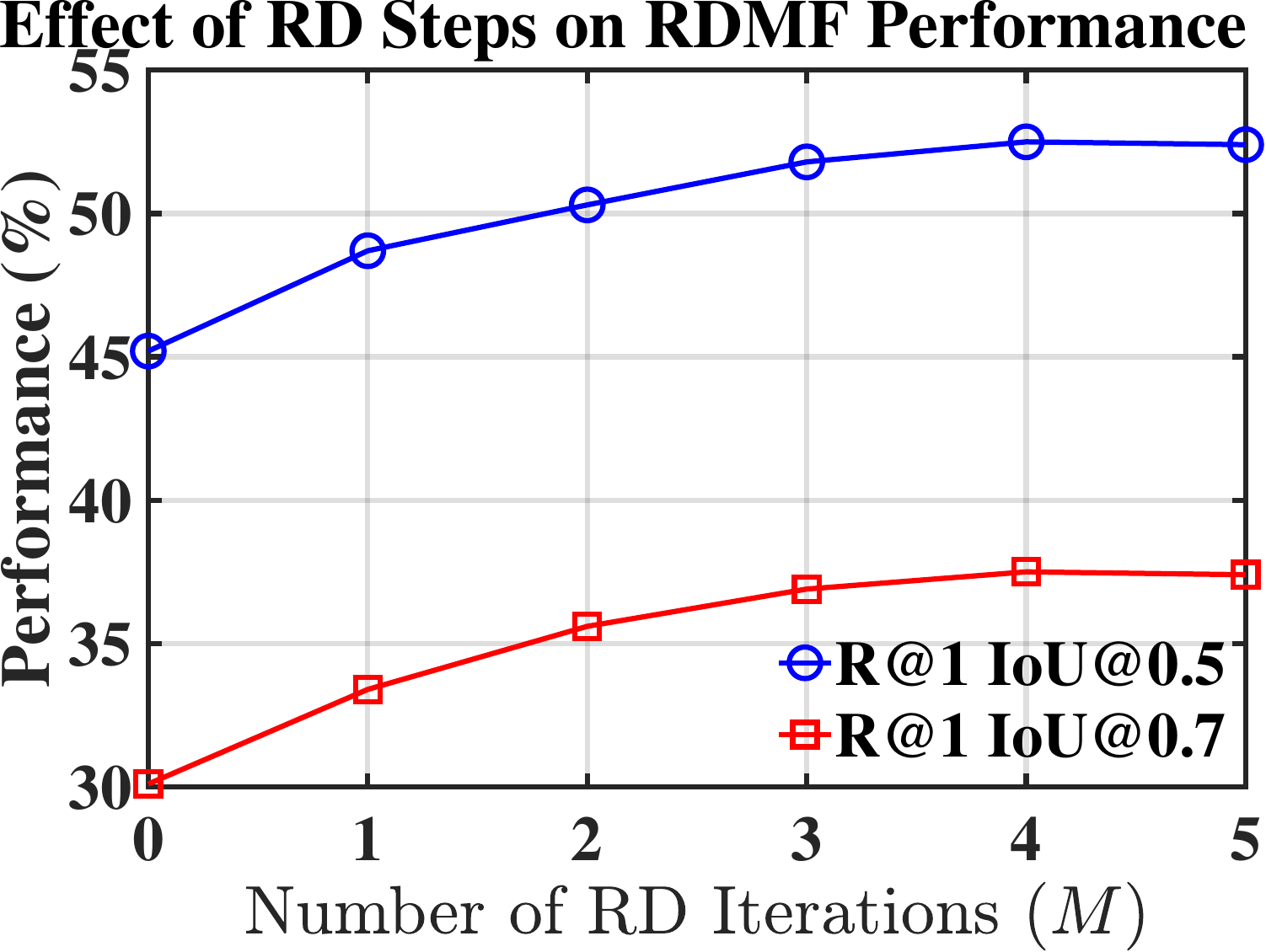} 
    \caption{Effect of the number of reaction-diffusion steps (\( M \))  on QVHighlights (left) and Charades-STA (right). }
    \vspace{-8pt}
    \label{fig:rd_steps}
\end{figure}

We also analyze the effect of the number of RD steps (\( M \)) in Figure \ref{fig:rd_steps}. Performance peaks at \( M=3 \), with diminishing returns beyond this point, suggesting that a few RD iterations are sufficient to form stable, meaningful patterns.

\subsubsection{Diffusion Coefficients (\( D_u, D_v \))}
The diffusion coefficients control the spatial spread of video (\(\mathbf{U}\)) and text (\(\mathbf{V}\)) features in the RD fusion module. Table \ref{tab:ablation_du_dv} shows performance for varying \( D_u \) and \( D_v \). The optimal configuration (\( D_u = 0.2, D_v = 0.1 \)) balances diffusion to enhance feature interactions without over-smoothing. Higher \( D_u \) (e.g., 0.3) reduces temporal specificity, while lower \( D_v \) (e.g., 0.05) limits text feature propagation, slightly degrading performance.

\begin{table}[t]
\centering
\caption{Ablation study on diffusion coefficients \( D_u \) and \( D_v \) on QVHighlights. Optimal performance is achieved at \( D_u = 0.2 \), \( D_v = 0.1 \).}
\vspace{-8pt}
\setlength{\tabcolsep}{6mm}{
\begin{tabular}{cccccccccc}
\toprule
\( D_u \) & \( D_v \) & {R@1@0.5 (\%)} & {AP (\%)} \\
\midrule
0.1 & 0.1 & 69.8 & 70.5 \\
0.2 & 0.1 & 70.5 & 71.2 \\
0.3 & 0.1 & 70.2 & 70.9 \\
0.2 & 0.05 & 70.1 & 71.0 \\
0.2 & 0.2 & 69.9 & 70.7 \\
\bottomrule
\end{tabular}}
\vspace{-8pt}
\label{tab:ablation_du_dv}
\end{table}

\subsubsection{Temporal Diffusion Steps (\( K \))}
Temporal diffusion smooths video features \(\mathbf{Z}_v\) before RD fusion (Equation \eqref{eq:diffusion}). Table \ref{tab:ablation_k} ablates the number of diffusion steps \( K \). Without diffusion (\( K=0 \)), performance drops due to noisy frame-level features. \( K=10 \) optimally smooths features, enhancing temporal coherence. Beyond \( K=15 \), over-smoothing reduces fine-grained details, slightly lowering scores.

\begin{table}[t]
\centering
\caption{Ablation study on temporal diffusion steps \( K \). \( K=10 \) yields the best performance.}
\vspace{-8pt}
\setlength{\tabcolsep}{9mm}{
\begin{tabular}{cccccc}
\toprule
\( K \) & {R@1@0.5 (\%)} & {AP (\%)} \\
\midrule
0 & 68.5 & 69.2 \\
5 & 69.7 & 70.4 \\
10 & 70.5 & 71.2 \\
15 & 70.3 & 71.0 \\
20 & 70.0 & 70.8 \\
\bottomrule
\end{tabular}}
\vspace{-8pt}
\label{tab:ablation_k}
\end{table}

\subsubsection{Reaction Function Parameters (\( F, k \))}
The reaction functions \( f, g \)  depend on feed rate \( F \) and kill rate \( k \). Table \ref{tab:ablation_f_k} shows their impact. The combination \( F=0.04, k=0.06 \) produces stable patterns (e.g., peaks at relevant frames). Lower \( F \) or higher \( k \) reduces pattern formation, while higher \( F \) or lower \( k \) causes instability, degrading performance.

\begin{table}[t]
\centering
\caption{Ablation study on reaction function parameters \( F \) and \( k \). Optimal values are \( F=0.04 \), \( k=0.06 \).}
\vspace{-8pt}
\setlength{\tabcolsep}{5mm}{
\begin{tabular}{ccccc}
\toprule
\( F \) & \( k \) & {R@1@0.5 (\%)} & {AP (\%)} \\
\midrule
0.02 & 0.06 & 69.5 & 70.2 \\
0.04 & 0.06 & 70.5 & 71.2 \\
0.06 & 0.06 & 70.1 & 70.8 \\
0.04 & 0.04 & 69.8 & 70.5 \\
0.04 & 0.08 & 69.6 & 70.3 \\
\bottomrule
\end{tabular}}
\vspace{-8pt}
\label{tab:ablation_f_k}
\end{table}

\subsection{Statistical Significance}
\label{app:stat_significance}

To validate RDMF’s superiority, we perform paired t-tests comparing R@1@0.5 and AP on QVHighlights. Table \ref{tab:stat_significance} reports p-values. All p-values are below 0.05, confirming that RDMF’s improvements are statistically significant. The smallest p-value against Moment-DETR (0.001) reflects RDMF’s substantial advantage over transformer-based methods, likely due to dynamic RD fusion.

\begin{table}[t]
\centering
\caption{Paired t-test p-values comparing RDMF with baselines on QVHighlights. All p-values \(< 0.05\), indicating significant improvements.}
\vspace{-8pt}
\setlength{\tabcolsep}{4mm}{
\begin{tabular}{lcc}
\toprule
Baseline & R@1@0.5 p-value & AP p-value \\
\midrule
CG-DETR & 0.002 & 0.003 \\
InternVideo & 0.008 & 0.007 \\
Moment-DETR & 0.001 & 0.002 \\
\bottomrule
\end{tabular}}
\vspace{-8pt}
\label{tab:stat_significance}
\end{table}

\subsection{Qualitative Analysis}
\begin{figure}[t]
    \centering
    \includegraphics[width=\columnwidth]{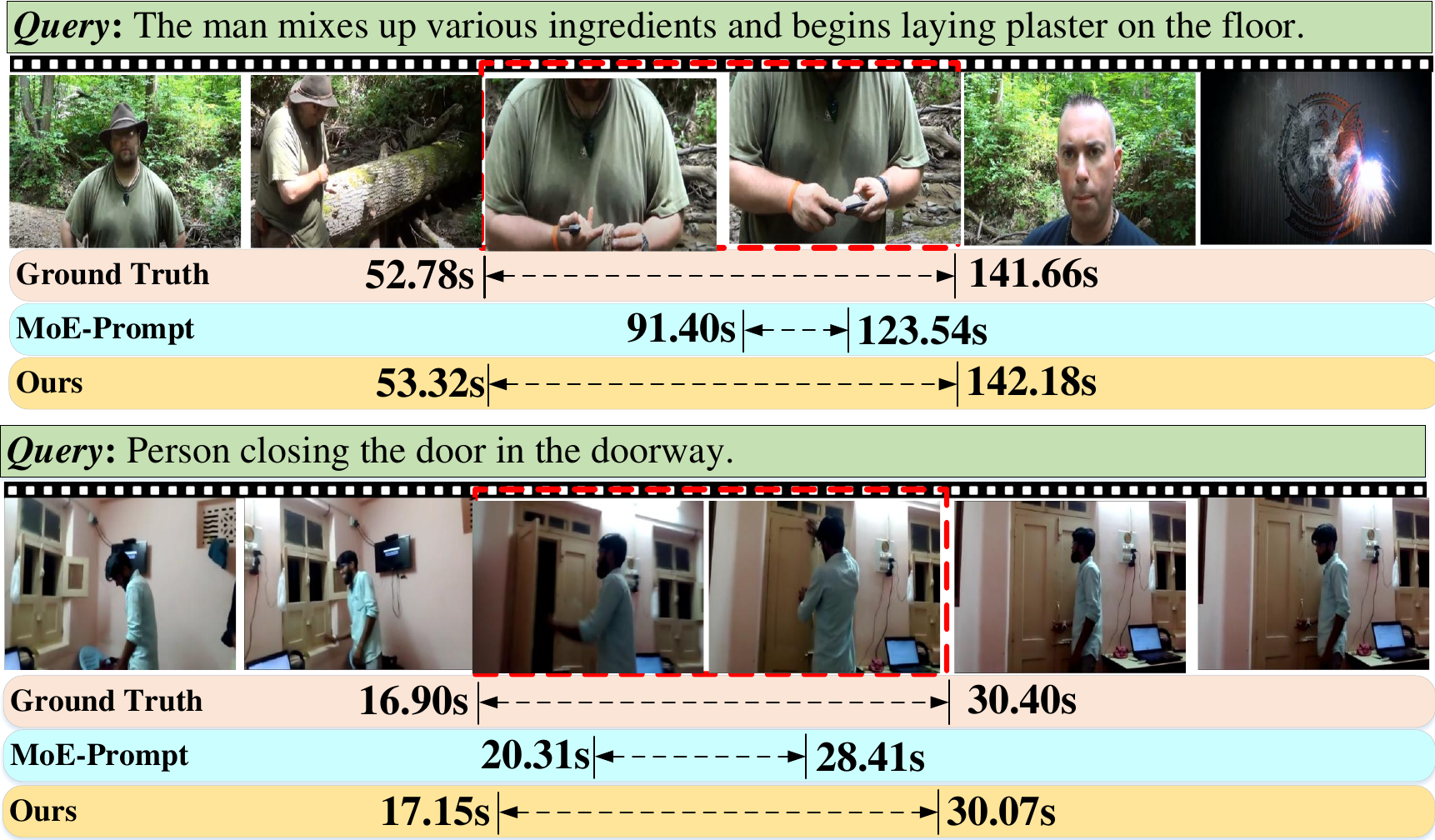}
    \caption{Qualitative comparison of saliency predictions on ActivityNet Captions (top) and Charades-STA (bottom). 
    }
    \vspace{-8pt}
    \label{fig:qualitative}
\end{figure}

Figure \ref{fig:qualitative} visualizes RDMF’s saliency predictions on ActivityNet Captions and Charades-STA. Our RDMF accurately highlights frames corresponding to the kicking action, with sharp temporal boundaries, while suppressing irrelevant background frames. In contrast, MoE-Prompt produces noisier saliency scores, misidentifying some non-kicking frames as salient. This demonstrates RDMF’s ability to form emergent patterns that align with semantic queries, a direct result of its RD-based fusion.

\subsection{Temporal Diffusion Effects}
\label{app:temporal_diffusion}

\begin{figure}[t]
\centering
\includegraphics[width=0.23\textwidth]{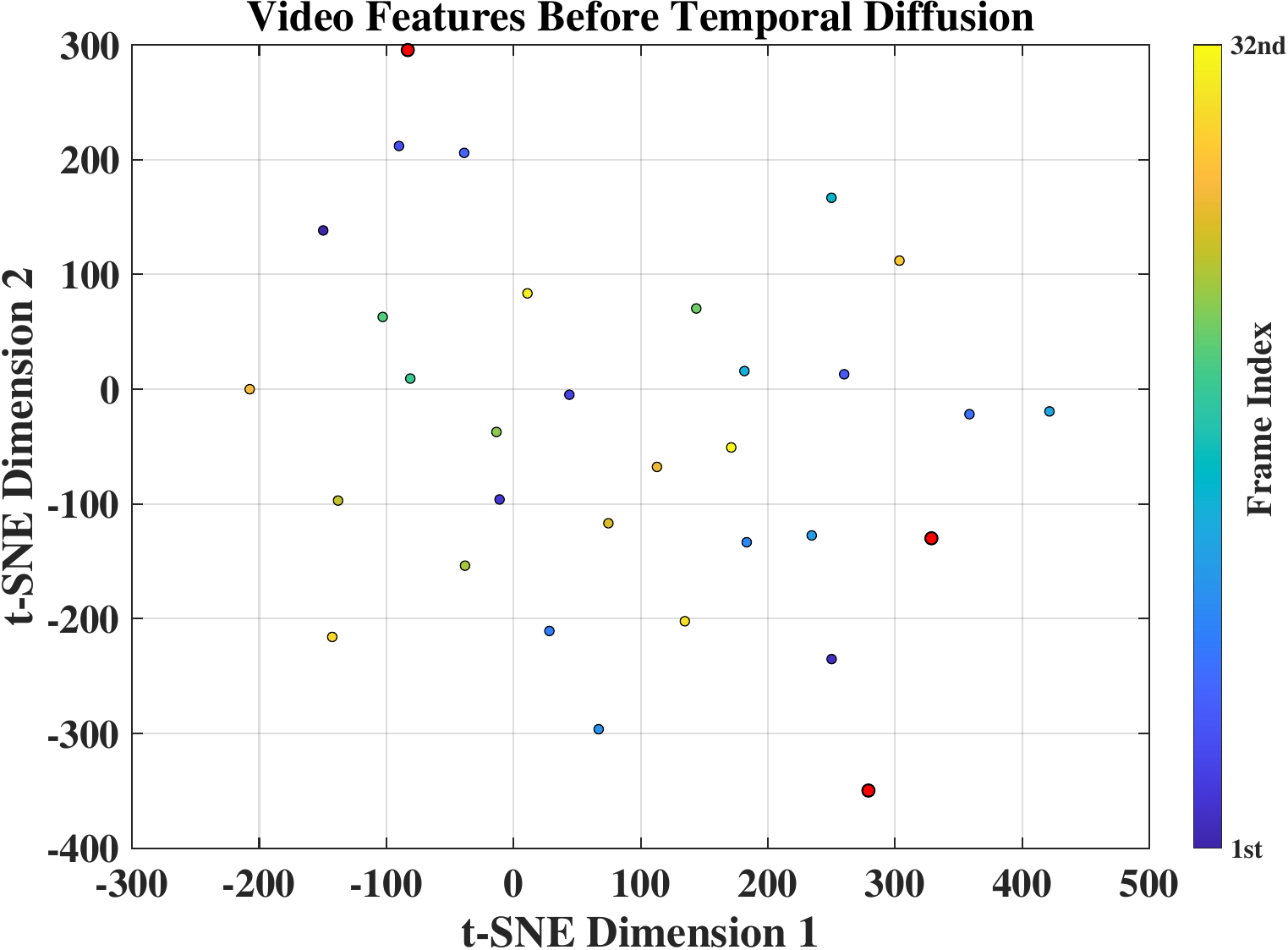}
\includegraphics[width=0.23\textwidth]{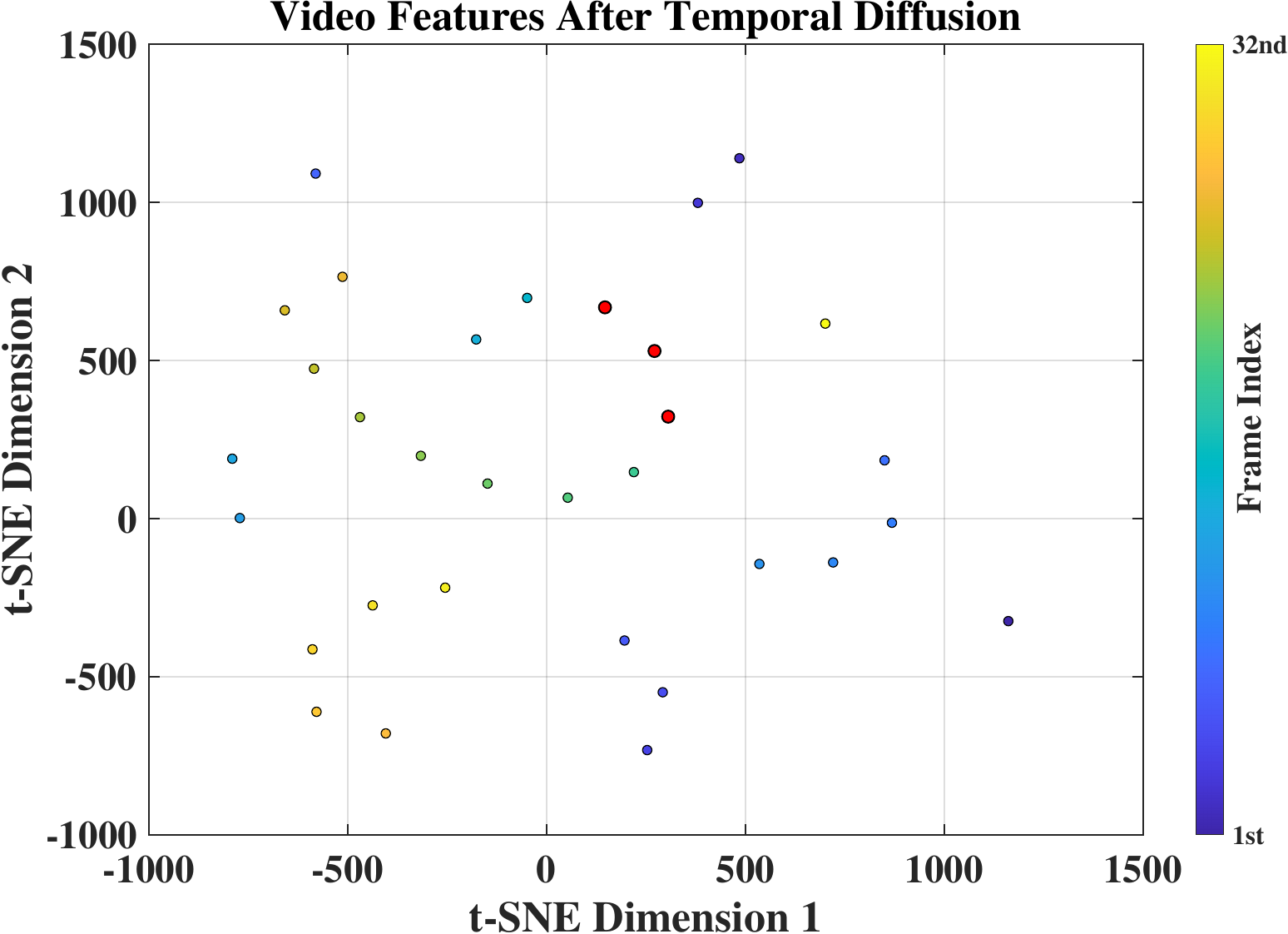}
\caption{Temporal diffusion effects on QVHighlights, where  Before diffusion (left): \(\mathbf{Z}_v^{(0)}\), raw video features; After diffusion (right): \(\mathbf{Z}_v^{(K)}\), smoothed features (\( K = 10 \)). Plots show feature values for a single dimension (index 100) across 32 frames. Diffusion reduces noise, enhancing temporal coherence around frames 16--18.}
\vspace{-8pt}
\label{fig:diffusion_effect}
\end{figure}

The temporal diffusion module smooths video features \(\mathbf{Z}_v\) over \( K = 10 \) steps to reduce noise. We visualize the effect by comparing \(\mathbf{Z}_v^{(0)}\) (raw features) and \(\mathbf{Z}_v^{(K)}\) (smoothed features) for a single feature dimension across frames. Figure \ref{fig:diffusion_effect} shows this for the same  video. The raw features \(\mathbf{Z}_v^{(0)}\)  exhibit high-frequency noise, with erratic fluctuations across frames. After \( K = 10 \) diffusion steps, \(\mathbf{Z}_v^{(K)}\) shows smoother variations, with a clear peak around frames 16--18, aligning with the jumping action. This smoothing, driven by the discrete Laplacian, improves robustness for RD fusion, as validated by ablation studies. The visualization highlights diffusion’s role as a low-pass filter, preserving relevant temporal patterns while suppressing noise.

\subsection{Discussion}
The experimental results validate RDMF’s contributions, confirming its ability to: 1)  \textbf{Capture Dynamic Interactions}: The RD mechanism outperforms static cross-attention and prompt-tuning methods by modeling non-linear video-text interactions, as evidenced by superior R@1 and AP scores. 2) \textbf{Leverage Interdisciplinary Insights}: The biologically inspired RD framework, grounded in the Gray-Scott model and Turing instability, enables adaptive pattern formation, setting RDMF apart from conventional methods. 3) \textbf{Enable Generalization}: RDMF’s performance on diverse datasets highlights its robustness across domains, unlike baselines that struggle with complex temporal relationships.

While RDMF shows promise, its computational complexity (due to multiple RD steps) is a limitation, which we address by optimizing the number of steps (\( M=3 \)). Future work could explore parallelized RD implementations or extensions to other multimodal tasks, such as audio-visual alignment. RDMF’s interdisciplinary approach opens new research avenues, encouraging the multimedia community to explore biologically inspired paradigms for multimedia challenges.

\section{Conclusion}
\label{sec:conclusion}

In this paper, we propose \textbf{Reaction-Diffusion Multimodal Fusion (RDMF)}, a novel framework for video-language modeling that reimagines multimodal fusion through the lens of systems biology. By drawing inspiration from reaction-diffusion (RD) systems, originally proposed by Alan Turing to explain pattern formation in chemical and biological processes, RDMF models the dynamic interplay between video and text modalities as a process of temporal diffusion and non-linear reactions. 
Our framework leverages the Gray-Scott RD model to enable adaptive pattern formation, amplifying text-aligned video features while suppressing irrelevant ones, resulting in superior performance for moment retrieval and highlight detection tasks. Our experiments on standard datasets,  demonstrate RDMF’s effectiveness, with significant improvements over state-of-the-art baselines.

Looking forward, several directions warrant exploration. First, optimizing the computational efficiency of the RD module, perhaps through parallelized implementations, could enable real-time deployment in resource-constrained environments. Second, extending RDMF to other multimodal tasks, such as video question answering or multimodal generation, could further validate its generalizability. Finally, integrating additional biological principles could inspire new paradigms for multimedia systems, fostering deeper collaboration between the multimedia community and fields like systems biology or computational neuroscience.

\bibliographystyle{ACM-Reference-Format}
\balance
\bibliography{main}

\end{document}